\documentclass[conference]{IEEEtran}
\IEEEoverridecommandlockouts
\usepackage{cite}
\usepackage{amsmath,amssymb,amsfonts}
\usepackage{algorithmic}
\usepackage{graphicx}
\usepackage{textcomp}
\usepackage{xcolor}

\usepackage{subfigure}

\usepackage{array}
\usepackage{float}
\usepackage{amsthm}
\usepackage{amssymb}
\usepackage{dblfloatfix}    % To enable figures at the bottom of page

\usepackage[ruled,vlined]{algorithm2e}

\usepackage{mathptmx}

\newtheorem{definition}{Definition}

\def\BibTeX{{\rm B\kern-.05em{\sc i\kern-.025em b}\kern-.08em
    T\kern-.1667em\lower.7ex\hbox{E}\kern-.125emX}}
\begin{document}

\title{imdpGAN: Generating Private and Specific Data with Generative Adversarial Networks}

% \author{Anonymous Authors}
\author{

\IEEEauthorblockN{Saurabh Gupta}
\IEEEauthorblockA{
\textit{IIIT - Delhi}\\
Delhi, India \\
saurabhg@iiitd.ac.in}

\and

\IEEEauthorblockN{Arun Balaji Buduru}
\IEEEauthorblockA{
\textit{IIIT - Delhi}\\
Delhi, India \\
arunb@iiitd.ac.in}

\and

\IEEEauthorblockN{Ponnurangam Kumaraguru}
\IEEEauthorblockA{
\textit{IIIT - Delhi}\\
Delhi, India \\
pk@iiitd.ac.in}

}

\maketitle

\begin{abstract}
Generative Adversarial Network (GAN) and its variants have shown promising results in generating synthetic data. However, the issues with GANs are: (i) the learning happens around the training samples and the model often ends up remembering them, consequently, compromising the privacy of individual samples - this becomes a major concern when GANs are applied to training data including personally identifiable information, (ii) the randomness in generated data - there is no control over the specificity of generated samples. To address these issues, we propose imdpGAN - an information maximizing differentially private Generative Adversarial Network. It is an end-to-end framework that simultaneously achieves privacy protection and learns latent representations. With experiments on MNIST dataset, we show that imdpGAN preserves the privacy of the individual data point, and learns latent codes to control the specificity of the generated samples. We perform binary classification on digit pairs to show the utility versus privacy trade-off. The classification accuracy decreases as we increase privacy levels in the framework. We also experimentally show that the training process of imdpGAN is stable but experience a 10-fold time increase as compared with other GAN frameworks. Finally, we extend imdpGAN framework to CelebA dataset to show how the privacy and learned representations can be used to control the specificity of the output.
\end{abstract}


\section{Introduction}
\noindent The world of today is moving towards more personalized hardware and software, collecting sensitive information with multiple Personally Identifiable Information (PII) attributes, especially in domains like healthcare and Internet-of-Things (IoT). Often times, deep learning techniques are used to solve problems like detecting cancer patterns \cite{cancers11091235}, diabetic retinopathy \cite{dr}, and so on. But deep learning typically needs huge amount of data to achieve promising performance. However, in domains like healthcare and IoT (with a lot of PII attributes), it is impossible to get as much data as we want. Also, such models learn finer details in training data and are shown to compromise privacy of individuals. One such example is successful recovery of individual samples from the training set, by using hill climbing on output probabilities \cite{Fredrikson:2015:MIA:2810103.2813677}. Therefore, enforcing privacy while using deep learning techniques to analyze such data has become an absolute necessity. In short there are two challenges: the availability of a huge amount of data and protecting the privacy of individual users.

% is a research by \cite{Fredrikson:2015:MIA:2810103.2813677}, who successfully recovered
% The data contains the highest degree of personal information about users. To improve such devices and their quality of service, organizations need to analyze the collected data to solve complex problems. At such a juncture, deep learning algorithms often seem to be the right choice while searching for a solution, but they require a lot of training samples. The scenario is commonplace, and the challenges are: protecting the privacy of these users and the availability of a huge amount of data.

Generative models have mitigated the data scarcity issue by successfully generating patient records, sensor data, medical records, tabular data  \cite{pmlr-v68-choi17a, Alzantot2017SenseGenAD, Camino2018GeneratingMS, Guan2018GenerationOS, Xu2018SynthesizingTD}. Using the combination of game theory and deep learning, GANs and its many other variants, have demonstrated promising performance in modeling the underlying data distribution \cite{NIPS2014_5423}. These generative models can generate high quality ``fake'' samples that are hard to differentiate from the real ones \cite{Mogren2016CRNNGANCR, Saito2016TemporalGA, Salimans2016ImprovedTF}. Ideally, we can generate these ``fake'' data samples to fit our needs and conduct the desired analysis without privacy implications. Although, the generation process is random and we cannot implicitly control the variation in type or style of data we want to generate. 

% %The fake samples and the learned distribution can be used for analysis while preserving privacy at the same time. 

Privacy is being enforced on sensitive data using several anonymization techniques. Some examples include k - anonymity \cite{Sweeney:2002:KAM:774544.774552}, l-diversity \cite{1617392}, t-closeness \cite{4221659}, which are effective but vulnerable to de-anonymization attacks \cite{Narayanan:2008:RDL:1397759.1398064}. Since, these techniques do not solve the data scarcity issue, researchers are trying to introduce privacy preservation in generative models \cite{Xie2018DifferentiallyPG}. The generation of ``fake'' samples is not self-sufficient and is prone to disclosure of private information about the individual training samples. The adversarial training procedure with high model complexity often leads to learning a distribution that just copies the training samples. Repeated sampling from such distributions increases the chance of recovering the training samples, hence, compromising the privacy of the data. \cite{Hitaj:2017:DMU:3133956.3134012} demonstrated an inference attack that uses generated samples to recreate the training samples. 

% %Therefore, generative models must be modified such that they not only generate high quality samples but protect the privacy of an individual training sample at the same time.  

% %Another problem with training GANs is mode-collapse. The generator collapses while training and produces only a limited variety of samples. Due to the mode-collapse, a GAN ends up generating an imbalanced dataset. 
% %\cite{Arjovsky2017WassersteinG} propose techniques to resolve the issues mentioned, and we leverage these techniques to build our model, to overcome these caveats.  

% %Privacy is being enforced on sensitive datasets using anonymization techniques. Some examples are k-anonymity \cite{Sweeney:2002:KAM:774544.774552}, l-diversity \cite{1617392}, t-closeness \cite{4221659}, and their variants, which are effective but vulnerable to de-anonymization attacks \cite{Narayanan:2008:RDL:1397759.1398064, Narayanan2006HowTB}. 

\textit{Our contribution.} With the above considerations, we try to learn meaningful latent representations of variation, known as latent codes, to have control over the specificity of the generator output and use a private training procedure to preserve the privacy of the individual training samples. Therefore, in this paper, we present an amalgamation of techniques from Information Maximizing Generative Adversarial Network, (used to learn interpretable latent representations in an unsupervised manner) and Differentially Private Generative Adversarial Network (used to preserve privacy of the training samples). The models are built using the machine learning framework Pytorch \cite{paszke2017automatic}. 

% Our main contributions are summarized as:

%\footnote{For reproducibility, we intend to make the source code of InfoDPGAN and the experiments publicly available upon publication.}. We summarize our contributions as follows:

We propose imdpGAN, a unified framework to: 
    \begin{enumerate}
        \item \textit{Protect privacy of training samples.}
        Protecting privacy in images simply means that one will not be able to recognize what is there in the image, i.e., the generator will generate blurry images as we increase privacy giving rise to a privacy versus accuracy trade-off. To demonstrate the trade-off, we train a binary classifier on digit pairs and find accuracy on corresponding test samples (discussed in Section \ref{sec:tradeoff}). Results show that as we increase privacy, the accuracy of binary classifier decreases.
        \item \textit{Control specificity of generated samples.}
        There are two kinds of variations in a data set: discrete and continuous. Discrete variations are represented by different classes. For example, MNIST dataset has 10 classes representing one digit per class. Changing one class to another is a discrete variation. The dataset has digits positioned at varying angles and having different widths representing continuous variation. We learn tunable latent codes to control both types of variations.  
    \end{enumerate}
    
We evaluate our proposed approach on MNIST dataset and extend the imdpGAN framework to complex CelebA \cite{liu2015faceattributes} dataset. Results show that imdpGAN preserves privacy and  learns meaningful latent codes, which are varied to show class and style variations while generating new images. Although, the classification accuracy decreases as we increase privacy. 

As privacy concerns are rising up there are multiple use cases of our framework. For example, popular face recognition systems (FRS) claim that they store only a representation of users' faces and not the actual image\footnote{Apple tweeted, ``Face ID only stores a mathematical representation of your face on iPhone, not a photo.'', https://twitter.com/apple/status/1215224753449066497}. However, while operating they require a complete face image as input to auhenticate an user. The proposed framework, imdpGAN, can be used to create anonymized face images that are closer to the real face representations by learning meaningful latent codes while generating private faces to preserve user's privacy. Although, there will be a trade-off between the accuracy of the FRS and privacy of the face image, which can further be adjusted by tunable parameters.

% We believe the framework has real-life applications and can be used to generate specific privacy-preserving synthetic samples in private data sharing scenarios.

% We want to create a model that gives a user tunable parameters to control the level of privacy and the kind of data they want in the generated samples.
% Figure 4 and Section 4.3 in the paper is precisely about the utility and privacy tradeoff. We show that - as we increase privacy, the classification accuracy (utility) decreases.
% We agree that MNIST dataset does not contain any private information. We chose MNIST/CelebA (publicly available datasets) to avoid revealing any private information in a publication. We ensure that with hyperparameter tuning, the framework can be used with other image datasets.
% The images getting blurred (noisy) show that generated images are getting more private, i.e., by looking at the images one can’t tell what digit they represent. Although, a reduction in classification accuracy can be seen due to the same phenomena. One can tweak the parameters based on usability to get desired images.

% \subsection{Organization}

This paper proceeds as follows: we start with a review of the background to explain the relevant work done on differential privacy for deep learning and learning latent representations in Section \ref{back}. In Section \ref{method}, after defining the privacy model and mutual information maximization used to learn latent codes, we introduce the imdpGAN framework followed by the differential privacy guarantees. In Section \ref{exp}, experiments on MNIST dataset are described, followed by the extended experiments on the CelebA dataset. Then we discuss the shortcomings of imdpGAN framework in Section \ref{disc} and finally conlcude in Section \ref{conc}.

\section{Background}
\label{back}
In this section, we provide a brief literature review of relevant topics: differential privacy for deep learning and learning disentangled representations.

\subsection{Differential Privacy for Deep Learning}
In the works that study differential privacy in deep learning, \cite{Abadi:2016:DLD:2976749.2978318} change the model's training algorithm to make it private by clipping and adding noise to the gradients. Authors also propose a privacy accounting technique and introduce a moments accountant that computes the privacy costs. In \cite{Shokri:2015:PDL:2810103.2813687}, authors use differential privacy with a parallel and asynchronous training procedure for a multi-party privacy-preserving neural network. It involves transmitting local parameters between server and local task, which has a high risk of information leakage. \cite{article111} models a private convolutional deep belief network by adding noise on its objective functions and an extra softmax layer. \cite{Xie2018DifferentiallyPG} leverages the moments accountant and the private training procedure from \cite{Abadi:2016:DLD:2976749.2978318} to train a differentially private generator. Authors add noise to the training procedure and avoid a distributed framework to prevent any information leaks. Advantages of DPGAN's techniques over other methods made them a salient choice for privacy preservation in the proposed framework \cite{Xie2018DifferentiallyPG}.

\subsection{Learning Latent Representations}
\label{llr}
Learning latent representations in a supervised, unsupervised, and semi-supervised manner is attempted by a lot of studies. Works that use supervision (labeled data): bilinear models \cite{Tenenbaum:2000:SSC:1121517.1121518} to separate style and content; multi-view perceptron \cite{Zhu:2014:MPD:2968826.2968851} to separate face identity from the viewpoint, train a subset of representation to match some supplied label using supervised learning. Then there are semi-supervised methods developed to eliminate the need for labels of variations. To disentangle representations, \cite{Reed:2014:LDF:3044805.3045052} proposes a higher-order Boltzmann machine, which uses a clamping technique. \cite{NIPS2015_5851} uses the clamping idea with variational autoencoders (VAEs) to learn codes that can represent pose and light in 3D rendered images. The model proposed by \cite{article112} can learn a representation that supports basic linear algebra on code space using GANs \cite{NIPS2014_5423}.
InfoGAN \cite{Chen:2016:IIR:3157096.3157340} learns disentangled representations with no supervision of any kind. Unlike hossRBM \cite{Desjardins_disentanglingfactors}, which can learn only the discrete latent factors and has high complexity, InfoGAN can learn both discrete and continuous latent representations and can scale to complicated data. We use the techniques from InfoGAN to learn factors of variation, called the latent codes, to control specificity in the proposed framework as it is easy to incorporate along with the privacy framework.

\section{Methodology}
\label{method}

We unify the techniques to stabilize the GAN training \cite{Arjovsky2017WassersteinG} with techniques to make a differentially private generator \cite{Xie2018DifferentiallyPG}, and learning latent representations \cite{Chen:2016:IIR:3157096.3157340}. In this paper, we propose an information maximizing differentially private Generative Adversarial Network (imdpGAN) that preserves privacy of the training samples and learns latent codes to control specificity of generated output. We discuss the privacy model in Section \ref{dpm}. Then we discuss the mutual information regularization used to learn factors of variation from the data in Section \ref{mi}. Then we explain imdpGAN framework, the objective function, and the private training procedure used to introduce differential privacy in Section \ref{frame}. Finally, we explain the privacy guarantees of imdpGAN in Section \ref{sec:guarantee}.

%We leverage it from  Differentially Private Generative Adversarial Network (DPGAN) \cite{Xie2018DifferentiallyPG}. DPGAN uses carefully designed noise and clipping to ensure differential privacy in the generator output. 
%We use the architecture from Information Maximizing Generative Adversarial Network (InfoGAN) \cite{Chen:2016:IIR:3157096.3157340} to learn interpretable latent representations. InfoGAN learns discrete and continuous latent codes that are used to control the specificity of the output.    

\begin{figure*}[!b]
    \centering
    \includegraphics[width=0.8\linewidth]{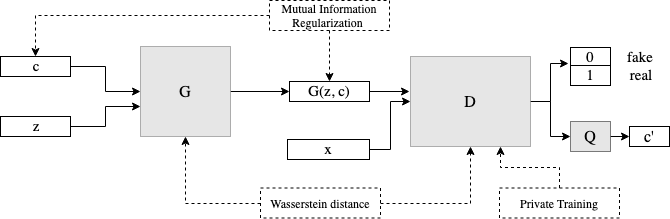}
    \caption{imdpGAN Architecture: addition of the private training procedure, the mutual information regularization and the Wasserstein distance.}
    \label{fig:arch}
\end{figure*}

\subsection{Differential Privacy}
\label{dpm}

We used differential privacy, as defined by \cite{Dwork:2006:DP:2097282.2097284}, as the privacy model for imdpGAN framework:
%Assume an algorithm, $A_p(.)$, having differential privacy property, is randomized such that it is difficult for an observer to backtrack to input-data using outputs of algorithm. Differential privacy (DP) is defined by:

\begin{definition}
\label{dp}
(Differential Privacy, DP) A randomized algorithm $A_{P}$ is ($\epsilon, \delta$)-differentially private if for any two databases $D$ and $D'$ differing in a single point and for any subset of outputs $S$:
\[ P(A_{P}(D) \: \in \: S) \; \leq \; e^{\epsilon} \:.\: P(A_{P}(D') \: \in \: S) \: + \: \delta \]
where $A_{P}(D)$ and $A_{P}(D')$ are the outputs of the algorithm for input databases $D$ and $D'$, respectively, and $P$ is the randomness of the noise in the algorithm.  
\end{definition}

%In \cite{Abadi:2016:DLD:2976749.2978318}, authors propose the moments accountant and the differentially private iterative gradient descent procedure that are the basis of introducing privacy and quantitatively measuring it in terms of $\epsilon$. 
\cite{Xie2018DifferentiallyPG} shows that definition in Theorem \ref{dp} is equivalent to:
\begin{displaymath}
     \left| \log\frac{P(A_{P}(D) \: = \: s)}{P(A_{P}(D') \: = \: s)}\right|  \; \leq \; \epsilon 
\end{displaymath}
with probability $1-\delta$ for every $s \in output$, where $\epsilon$ is the privacy level. A small $\epsilon$ value indicates the $A_{P}$'s output probabilities differ by a small value at $s$ indicating high fluctuations of ground truth outputs and hence high privacy. On the other hand, $\epsilon=\infty$ means no noise or simply the non-private case.

According to Definition \ref{dp}, and the above intuition, the $\epsilon$ values represent what level of privacy is protected of individual sample from the dataset. For example, when collecting sensitive information for some experiment, sometimes an individual does not want an observer to know their involvement in the experiment. This is because then the observer can harm that individual's interest. Preserving this involvement would ensure the protection of individual's privacy. It will also make sure that the result will not affect too much if we replace this individual with someone else, which is what we plan to achieve using differential privacy.

\subsection{Mutual Information Regularization}
\label{mi}
    In traditional GANs \cite{NIPS2014_5423}, the generator uses a simple input noise vector $z$ and imposes no constraints on how the noise is used. As a result, noise is used in a highly entangled way and prevents from learning mappings of $z$ to semantic features in the data. The results from the generator of such GANs is, therefore, highly random. However, many datasets decompose into a set of meaningful factors of variation. For example, the MNIST dataset has ten classes, and within the dataset, the thickness and angle of the digits vary. Ideally, a model should automatically learn a discrete random variable to represent the classes and a continuous random variable to represent the thickness and angle properties for such dataset.
    
    To target the structured semantic features of data distribution, \cite{Chen:2016:IIR:3157096.3157340} decomposes the input noise vector into two parts: the noise vector $z$ and the latent code $c$. The output of the generator becomes $G(z,c)$. In traditional training, the generator can freely ignore the additional latent code, $c$. Therefore, to introduce dependency between $c$ and $G(z,c)$, a mutual information term, $I(c;G(z,c))$ is used as a regularization term given as:
    
    \begin{equation}
    \label{eq:i}
        \begin{aligned}
        I(c;G(z,c)) = H(c) - H(c|G(z,c)) \\ = H(G(z,c)) - H(G(z,c)|c)
        \end{aligned}
    \end{equation}

    where $H(c|G(z,c))$, $H(G(z,c)|c)$ are conditional entropies and $H(c)$, $H(G(z,c))$ are entropies.
    
    The regularization ensures high mutual information between latent code $c$ and generator output $G(z,c)$. It helps the model learn meaningful latent representations. The mutual information term maximizes the dependency between the latent codes and generator output. If $c$ and $G(z,c)$ are independent, then the term $I(c; G(z,c))$ becomes zero, i.e., one variable does not reveal anything about the other. In contrast, to maximize $I(c; G(z,c))$, we relate $c$ and $G(z,c)$ using a non-linear mapping. In the framework, the non-linear mapping is realized using a neural network, $Q$. The term $I(c;G(z,c))$ is hard to maximize directly, therefore, we define a variational lower bound, $L_{I}(G,Q)$ to maximize the mutual information \cite{Barber2003TheIA}.

\subsection{imdpGAN framework}
\label{frame}

We started from a basic GAN architecture but replaced the K-L, J-S divergence with Wasserstein distance as it makes the training stable \cite{Arjovsky2017WassersteinG}. We passed an additional input, i.e. the latent code, to the generator and added a mutual information regularization term to introduce dependency between the latent code and the generator output to learn meaningful latent representations.

\subsubsection{\textbf{Objective Function}}
As shown in Figure \ref{fig:arch}, we passed a noise vector $z$ and a latent code vector $c$ as input to the generator, $G$. The output from the generator, $G(z,c)$, and the real samples, $x$, were given to the discriminator, $D$. Recall that the objective function of a traditional GAN is:

\begin{equation}
\label{eq:1}
  \begin{aligned}
    \underset{G}{min}\:\underset{D}{max}\:V(D,G) \;=\; \mathbb E_{x \sim P_{data}}[\log D(x)] \\
        \; + \; \mathbb E_{z \sim P_{noise}}[\log (1 - D(G(z)))]
    \end{aligned}
\end{equation}

When we use Wasserstein distance \cite{Arjovsky2017WassersteinG} instead of the K-L, J-S divergence, the objective function in equation \ref{eq:1} changes as:

\begin{equation}
\label{eq:2}
    \begin{aligned}
        \underset{G}{min} \: \underset{w \in W}{max} \mathbb E_{x \sim P_{data}}[f_{w}(x)] \; - \;  \mathbb E_{z \sim P_{noise}}[f_{w}(G(z))]
    \end{aligned}
\end{equation}

where functions $ { f_{w}(x) }_{w \in W} $ are K-Lipschitz, which is a condition required to solve equation \ref{eq:2}.

Further, when we incorporated the mutual information regularization term and the auxiliary distribution Q, the objective function in equation \ref{eq:2} becomes:

\begin{equation}
\label{eq:3}
    \begin{aligned}
        V_{imdpGAN}(f_{w},G,Q) \;=\; \underset{G,Q}{min} \: \underset{w \in W}{max} \mathbb E_{x \sim P_{data}}[f_{w}(x)] \\ \; - \;  \mathbb E_{z \sim P_{noise}}[f_{w}(G(z))] - \lambda L_{I}(G,Q)
    \end{aligned}
\end{equation}

where $L_{I}(G, Q)$ is variational lower bound used to optimize the mutual information term. $\lambda$ is an extra hyperparameter used to scale the mutual information according to the GAN objectives.

From Figure \ref{fig:arch}, $D$ is trained using the private procedure explained in next section. $Q$ is trained in a manner to maximize the mutual information. The learned latent code, $c'$, can be used to control the specificity of the output.

\subsubsection{\textbf{Private training Procedure}}
\label{sec:ptp}
\cite{Abadi:2016:DLD:2976749.2978318} describes a differentially private training procedure for stochastic gradient descent that involves adding noise to the gradients and clipping the parameters of the discriminator. \cite{Xie2018DifferentiallyPG} extended the procedure to DPGANs. We use the extended version to formulate a private training procedure. The private procedure used in imdpGAN is summarized in Algorithm \ref{privateproc}. Adding noise to each gradient step ensures local differential privacy. We get a differentially private generator at the end of the training.

\begin{algorithm}
\SetAlgoLined
 \KwIn{Noise - $z$, real samples - $x$, parameter clip constant - $c_{p}$, batch size - $m$, total number of training samples - $M$, number of generator and discriminator iterations - $n_{g}$ and $n_{d}$, weights of generator and discriminator - $w_{g}$ and $w_{d}$, noise - $N$, noise scale - $\sigma$.}
 \KwOut{Differentially private $Generator$}
 \For{$ Generator \: Iterations, n_g $}{
    \For{ $ Discriminator\: Iterations, n_d $ }{
        
        Sample $ \left\{ z^{(i)} \right\}_{i=1}^m \sim p(z)$ a batch of samples \;
        Sample $ \left\{ x^{(i)} \right\}_{i=1}^m \sim p_{data}(x)$ a batch of real data points \;
        
        For each sample i, compute gradient of WGAN - $g_{w_{d}}(x^{(i)}, z^{(i)})$ \; 
        \textbf{Add noise to the gradient} \;
         $g_{w_d} \gets g_{w_d} + N(0, \sigma_n^2 c_p^2 I)  $  \;
         Update discriminator weights, $w_{d}$ \;
        \textbf{Clip the parameters} \;
        $w_{d} \gets clip(w_{d}, -c_{p}, +c_{p}) $ \;
    }
    Sample $ \left\{ z^{(i)} \right\}_{i=1}^m \sim p(z)$ another batch of samples \;
    Update generator gradient, $g_{w_{g}}$ and weights, $w_g$ \;
 }
 \textbf{return} $Generator$ \;
 \caption{Private Training Procedure}
 \label{privateproc}
\end{algorithm}

\subsection{Privacy Guarantees of imdpGAN}
\label{sec:guarantee}

To prove the privacy guarantees of imdpGAN, we show that the output of generator (through parameters of discriminator, $w_d$) guarantees differential privacy with respect to the training samples. Therefore, no generated output from G will compromise the privacy of training points. We can compute the final privacy value, $\epsilon$, using the moments accountant mechanism. 

Assume two generator iterations $t_1$ and $t_2$. By treating the parameters of discriminator, $w_d$ at $t_1$ as one point in outer space, it can be seen that the procedure to update $w_d$ from Algorithm \ref{privateproc} for fixed $t_2$ is just the algorithm $A_p$ in Definition \ref{dp}. So, we have $A_p(D) = M(aux, D)$, where $aux$ is just auxiliary input, which refers to $w_d$ at iteration $t_1$. On combining with Definition \ref{dp}, the privacy loss at point $o$ can be defined as:

\begin{definition}{Privacy Loss}
\begin{displaymath}
      c(o; M, aux, D, D') \triangleq log\frac{Pr[M(aux, D) = o]}{Pr[M(aux, D')= o]} 
\end{displaymath}
\end{definition}

% which is the difference calculated between two distributions due to the difference of samples present in $D$ and $D'$. The random variable that defines privacy loss at $o$ is:
% $C(M, aux, D, D') = c(M(D); M, aux, D, D')$
% defined by evaluating the privacy loss at a sample outcome from $M(D)$. Note that we assume the supports of two distributions 

The state of discriminator weights is updated by sequentially applying differentially private mechanisms. This is an instance of \textit{adaptive mechanism} modelled by letting $aux$ of $k^{th}$ mechanism, $M_k$, to be the output of all previous mechanisms. For a given mechanism, M, we define $\lambda^{th}$ moment $\alpha_{M} (\lambda; aux, D, D')$ as the log of moment generating function evaluated at $\lambda$: 

\begin{definition}
{Log moment generating function}
\begin{displaymath}
      \alpha_{M} (\lambda; aux, D, D') \triangleq \log \mathbb{E}_{o \sim M(aux, D)}[\exp(\lambda C(M, aux, D, D'))] 
\end{displaymath}
\end{definition}

The worst case scenario of moment generating function, known as the moments accountant, can be written as:

\begin{definition}{Moments accountant}
\begin{displaymath}
      \alpha_{M} (\lambda) \triangleq \underset{aux, D, D'}{\max} \alpha_{M} (\lambda; aux, D, D') 
\end{displaymath}
\end{definition}

The definition of moments accountant has properties as explained in \cite{Abadi:2016:DLD:2976749.2978318} (Theorem 2): i) composability - the overall moments accountant can be bounded by the sum of moments accountant in each iteration, i.e, privacy is proportional to iterations, ii) the tail bound can also be applied in privacy guarantee.

We need $g_{w_d}(x^{(i)}, z^{(i)})$ to be bounded (by clipping the norm and adding noise according to this bound in Algorithm \ref{privateproc}) to use the moments accountant. \cite{Xie2018DifferentiallyPG} (Lemma 3.5) proposes that by only clipping on $w_d$, we can automatically guarantee a bound on $g_{w_d}(x^{(i)}, z^{(i)})$. The lemma is given as:

\begin{definition}
\label{lemma:1}
Under the condition of Algorithm \ref{privateproc}, assume that the activation function of the discriminator has a bounded range and bounded derivatives everywhere: $ \sigma(.) \leq B_\sigma $ and $\sigma(.) \leq B_{\sigma'}$, and every data point $||x|| \leq B_x$, then 
$|| g_{w_d}(x^{(i)}, z^{(i)})||$ $\leq$ $c_p$ for some constant $c_p$.
\end{definition}

\cite{Xie2018DifferentiallyPG} proves that the Definition \ref{lemma:1} holds true if the following condition on derivatives of objective function is met:

\begin{equation}
\label{eq:der}
    \centering
    \begin{aligned}
          ||g_{w_d}(x^{(i)}, z^{(i)})|| = \left|\left| \nabla_w \left( f_w(x^{(i)}) - f_w(g(z^{(i)})) \right) \right|\right| \\
          \leq 2 \left|\left| \nabla_w f_w (x^{(i)}) \right|\right|
    \end{aligned}
\end{equation}

The derivative of our objective function, defined in Equation \ref{eq:3} is: 
\begin{displaymath}
    \begin{aligned}
        \left|\left| \nabla_w \left( f_w(x^{(i)}) - f_w(g(z^{(i)})) - \lambda L_{I}(g(z^{(i)}, c^{(i)}),Q) \right)  \right|\right| 
    \end{aligned}
\end{displaymath}

Since the additional mutual information regularization term, $\lambda L_{I}(g(z^{(i)}, c^{(i)}),Q)$ is always positive, the condition in Equation \ref{eq:der} holds true for this objective function as well. Therefore, in Algorithm \ref{privateproc}, we only clip $w_d$ to guarantee a bound on $g_{w_d}(x^{(i)}, z^{(i)})$.

Using Definition \ref{lemma:1}, \cite{Xie2018DifferentiallyPG} (Lemma 1.) proves the following Definition \ref{l2}. The definition holds true for our objective function as well,and therefore, guarantees differential privacy for discriminator training procedure.

\begin{definition}
\label{l2}
Given the sampling probability $q = \frac{m}{M}$ with $M$ as total number of samples and $m$ as batch size, the number of discriminator iterations in each inner loop $n_d$ and privacy violation $\delta$, for any positive $\epsilon$, the parameters of discriminator guarantee $(\epsilon, \delta)$-differential privacy with respect to all the data points used in that outer loop, $generator \: Iterations, n_{g}$ if we choose:
\begin{equation}
\label{sigep}
     \sigma \;=\; \frac{2q\sqrt{ n_d \log\frac{1}{\delta} }}{\epsilon}  
\end{equation}
\end{definition}

% To compute the overall privacy cost of the training, \cite{Abadi:2016:DLD:2976749.2978318} invented a \textbf{moments accountant} that computes the privacy cost whenever training samples are accessed. \cite{Xie2018DifferentiallyPG} changes the moments accountant to give a relationship between the noise, $\sigma$, and a privacy level, $\epsilon$ as:
% From Algorithm \ref{privateproc}, for a  given sampling probability $q = \frac{m}{M}$, 

Equation \ref{sigep} quantifies the relationship between privacy level $\epsilon$ and noise level $\sigma$. Tuning the noise level is required to have a reasonable privacy level. Note that for a fixed value of $\sigma$, larger $q$ will lead to less privacy guarantee because when more samples are involved less privacy is assigned on each of them. Also, as the iterations ($n_d$) increase less privacy is guaranteed because the training reveals more information about the data (specifically, more accurate gradients). Therefore, there is a need to choose a reasonable privacy level.

To achieve different levels of privacy, we use a noise distribution with zero mean and varying standard deviation.

\section{Experiments}
\label{exp}

\subsection{Experimental Setup} 
\label{setup}

The training of traditional GANs \cite{NIPS2014_5423} has problems with vanishing gradient and stability. \cite{Arjovsky2017WassersteinG} proposed the use of Wasserstein distance between the generator and the data distribution replacing the loss function such that a non-zero gradient always exists.
We used the same network architecture given by \cite{Chen:2016:IIR:3157096.3157340}. However, we added the private training procedure and used Wasserstein distance for training. Note that the mutual information regularization was already available with the base architecture.  

% % \cite{NIPS2015_5851} propose DCGAN architecture and show promising results in image generation using GANs. We used the same generator and discriminator as used in DCGAN architecture. Moreover, the training of traditional GANs \cite{NIPS2014_5423} has problems with vanishing gradient and stability. \cite{Arjovsky2017WassersteinG} explains these challenges and make use of Wasserstein distance between the generator and the data distribution replacing the loss function such that a non-zero gradient always exists. 

\subsubsection{\textbf{Introducing Privacy} }
We make the training procedure private by adding noise to discriminator gradients and clipping parameters of the discriminator as explained in Algorithm \ref{privateproc}. Instead of adding noise on the final output directly (global differential privacy), we focused on preserving privacy during the training (local differential privacy) as it generally results in high utility. We used Gaussian noise with zero mean (no bias) and varying standard deviation, $\sigma$. Gaussian noise is a popular choice for privacy preservation \cite{Dwork:2014:AFD:2693052.2693053} and usually results in $(\epsilon, \delta)$-differential privacy. From equation \ref{sigep}, $\sigma$ and $\epsilon$ are inversely corelated. Therefore, more noise results in smaller $\epsilon$ values. The batch size for all experiments is 64, and the number of samples is 60,000 and 202,599 for MNIST and CelebA, respectively. We set the sampling probability to $q = 64/number\_of\_samples$, the clip constant ($c_p$) to $0.01$ such that weights of discriminator are clipped back to $[-c_p, c_p]$, privacy violation ($\delta$) to $10^{-5}$, and number of discriminator iterations ($n_d$) to 5.

\subsubsection{ \textbf{Learning Latent Representations} }  The term $I(c;G(z,c))$, from equation \ref{eq:i}, is hard to maximize directly. Therefore, we defined an auxiliary distribution, $Q$, and used variational information maximizing term,  $L_{I}(G,Q)$ \cite{Barber2003TheIA} to maximize the mutual information. In imdpGAN, we formulated Q as a neural network. Q and the discriminator share all convolutional layers. To learn latent representations, we modeled discrete codes with uniform categorical distribution, $Cat(K=k, p=k/100)$ that model the discontinuous variation in data, i.e., classes. We also modeled continuous codes with a uniform continuous distribution, $Unif(-1,1)$ that capture continuous variations in data, like writing style (width) in MNIST, color variation in CelebA, etc.

\begin{figure*} [htbp]
\centering
  \subfigure[$\epsilon=\infty$]{
    \includegraphics[width=0.40\linewidth]{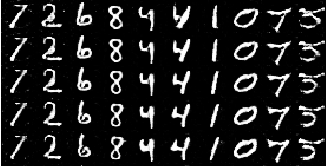}
    \label{fig7:a} 
  }
  \subfigure[$\epsilon=5.5$]{
    \includegraphics[width=0.40\linewidth]{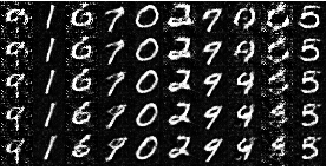}
    \label{fig7:b} 
  } 
  \subfigure[$\epsilon=2.2$]{
    \includegraphics[width=0.40\linewidth]{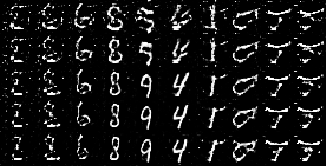}
    \label{fig7:c} 
  }
  \subfigure[$\epsilon=1.22$]{
    \includegraphics[width=0.40\linewidth]{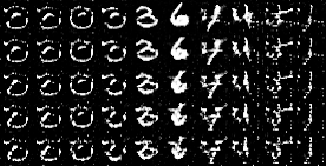}
    \label{fig7:d} 
  } 
  \caption{Varying latent codes to generate images on MNIST dataset. In all the images, the categorical discrete latent code is varied from left to right. For different $\epsilon$ values the change in the discrete latent code can be seen as the class of generated image is changing. The continuous latent code $c2$ is varied from -1 to +1 (top to bottom). A change in width of the generated samples can be observed with varying $c2$.}
  \label{fig7} 
\end{figure*}

\begin{figure*}[htbp]
  \subfigure[$\epsilon=\infty$]{
    \centering
    \includegraphics[width=0.23\linewidth]{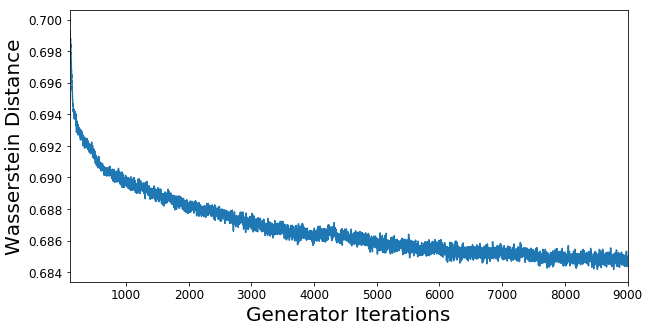}
    \label{fig2:a} 
  }
  \subfigure[$\epsilon=5.5$]{
    \centering
    \includegraphics[width=0.23\linewidth]{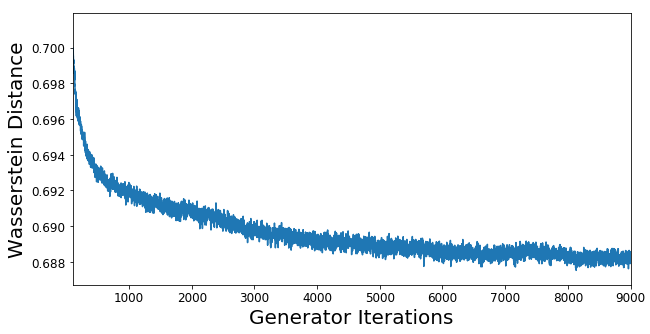}
    \label{fig2:b} 
  } 
  \subfigure[$\epsilon=2.2$]{
    \centering
    \includegraphics[width=0.23\linewidth]{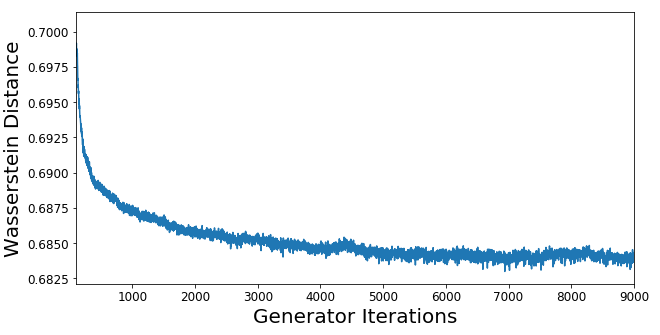}
    \label{fig2:c} 
  }
  \subfigure[$\epsilon=1.22$]{
    \centering
    \includegraphics[width=0.23\linewidth]{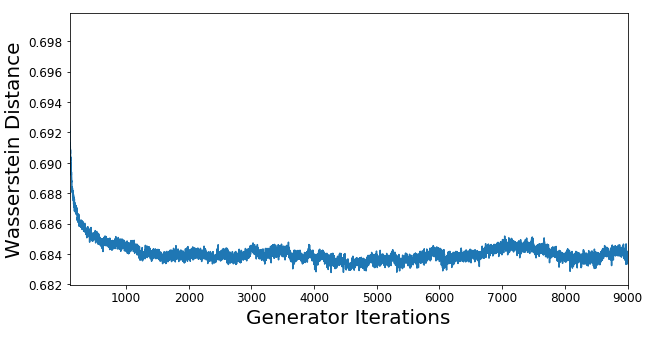}
    \label{fig2:d} 
  } 
  \caption{Wasserstein distance versus generator iterations on MNIST dataset. As $\epsilon$ values decrease, curves exhibit more fluctuations (due to more noise being added) and larger variance but converge proving training stability.}
  \label{fig2} 
\end{figure*}

\subsection{ imdpGAN Generation }
\label{nnn}

In this experiment, we study the effect of noise\footnote{Noise here refers to the Gaussian noise we add to the training procedure to introduce local differential privacy and not the noise vector $z$ we pass as input to the generator.} on the generator samples. We use MNIST data with 60,000 training samples to train imdpGAN. The experiment is run several times with varying privacy levels, $ 1 \leq \epsilon \leq 10$. The samples generated with $\epsilon \geq 10$ looked similar to the samples generated with $\epsilon = \infty$. To show the results we pick three $\epsilon$ values, $1.22$, $2.2$ and $5.5$ as the samples generated with these values looked significantly different. 

We wanted to see the output change as we make changes to the latent codes. The experiment is performed on MNIST dataset, with 60,000 training samples, 10 classes, and continuous variations like the rotation and width of the digit. We use a discrete-valued latent code, $c1 \sim Cat(K=10, p=0.1)$, which models the discontinuous variation in data, i.e., classes and a continuous latent code, $c2 \sim Unif(-1,1)$, which captures continuous variation in data such as the writing style. The generator is differentially private, so the samples generated with the model trained with smaller $\epsilon$ values (more noise added during training) are highly distorted. Due to distortion, the specificity in such images is not clearly visible. Although the learned representations are evident in at least one or more columns, as shown in Figure \ref{fig7}.

\subsubsection{\textbf{Convergence of network}}

The experiment is performed on MNIST dataset with 60,000 training samples to see the behavior of Wasserstein distance. We plot the Wasserstein distance versus generator iterations for every batch in MNIST dataset. The min-max training causes some fluctuations itself; therefore, to perform this experiment, only the amount of noise is changed, keeping the rest of the parameters fixed. As shown in Figure \ref{fig2}, the Wasserstein distance decreases during the training and converges in the end. We observed that a smaller $\epsilon$, indicating more noise, leads to more fluctuations and larger variance. The trend is visible in the later half of the plot. Intuitively, more noise should result in more fluctuations and hence, blurry images, which shows the experiment's consistency with the results of the previous experiment.
% % Differentiate between noise added and GAN noise

\subsubsection{\textbf{Binary Classification on Digit Pairs}}
\label{sec:tradeoff}

\begin{figure}[htbp]
    \centering
    \includegraphics[width=0.8\linewidth]{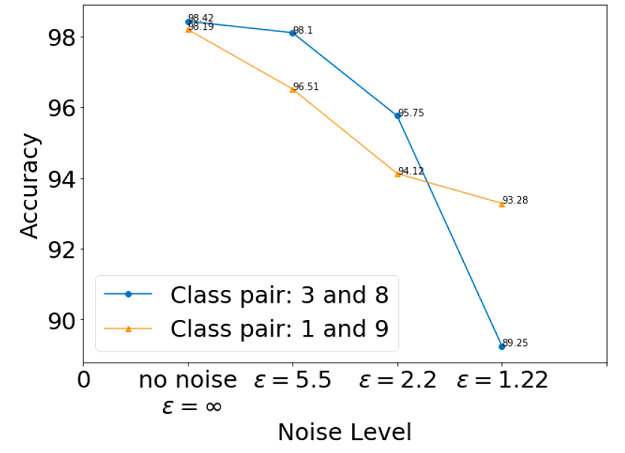}
    \caption{Binary Classification task on MNIST test set using samples generated with different $\epsilon$ values as training set. From left to right, we use generated data $\epsilon=\infty$ (without noise), and generated data with values $\epsilon = 5.5, 2.2, 1.22$ (with noise). We can see as more noise is added, i.e., more privacy is introduced, the accuracy of classifier decreases, which indicates a trade-off between choosing a promising privacy level and an acceptable accuracy threshold.}
    \label{fig:classifiers}
\end{figure}

In this experiment, we use a binary classification task to demonstrate the trade-off between utility and privacy of the framework. The classification acts as a quantitative measure to evaluate the utility of proposed framework. Since the images are blurred due to private training, we choose the classes which are still somewhat recognizable from all cases of $\epsilon$. We make pairs of these, viz. 3-8, 9-1, to perform binary classification. We choose 2,000 samples for each class for the no noise case and for $\epsilon = 5.5, 2.2, 1.22$. Then we build binary classifiers using the 4,000 samples (2,000 of each class) for class pairs 3-8 and 9-1. We perform the training for 100 epochs and report the accuracy of built classifiers on MNIST's test set. The results are shown in Figure \ref{fig:classifiers}. As expected, as the privacy level ($\epsilon$) increases, the accuracy of the classifiers decrease.

\subsubsection{\textbf{Mutual Information Maximization}}

\begin{figure}[htbp]
    \centering
    \includegraphics[width=0.8\linewidth]{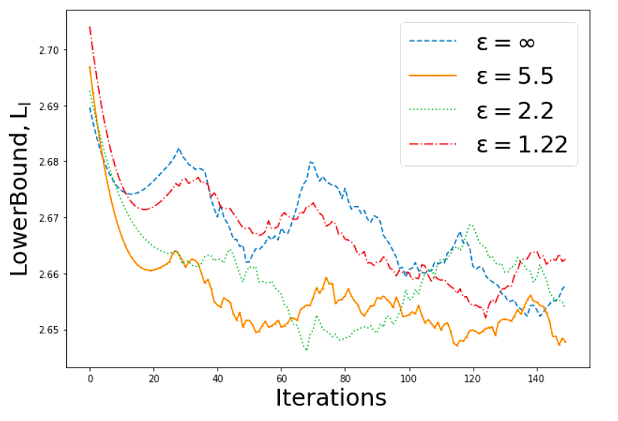}
    \caption{The Lower Bound, $L_{I}$, versus Iterations on MNIST dataset. The lower bound decreases with decreasing $\epsilon$ values. The plot demonstrates that $\epsilon$ values must be chosen carefully to learn meaningful latent representations.}
    \label{fig:mi}
\end{figure}

The experiment demonstrates the changes in the mutual information, between the generator and the latent codes, with changing privacy levels. We trained imdpGAN on MNIST dataset with 60,000 training samples. We defined the latent code using a categorical uniform distribution, $c \sim Cat(K=10, p=0.1)$. We trained imdpGAN with different privacy levels. As shown in Figure \ref{fig:mi}, as we add more noise (decreasing $\epsilon$ values), the mutual information gain decreases. The pattern demonstrates that the lower bound, $L_{I}$, is quickly maximized but due to the addition of noise at each iteration, its value decreases. Therefore, noise must be chosen very carefully to maintain a good value of $L_{I}$, which ensures the learning of meaningful latent representations.

% \subsection{E2: Controlling the Specificity of Output using Latent Codes}

% %To learn latent representations for MNIST dataset, 
% We wanted to see the output change as we make changes to the latent codes. The experiment is performed on MNIST dataset, with 60,000 training samples, 10 classes, and continuous variations like the rotation and width of the digit. We use a discrete-valued latent code, $c1 \sim Cat(K=10, p=0.1)$, which models the discontinuous variation in data, i.e., classes and a continuous latent code, $c2 \sim Unif(-1,1)$, which captures continuous variation in data such as the writing style. The generator is differentially private, so the samples generated with the model trained with smaller $\epsilon$ values (more noise added during training) are highly distorted. Due to distortion, the specificity in such images is not clearly visible. Although the learned representations are evident in at least one or more columns, as shown in Figure \ref{fig7}. 

% \begin{figure*}[htbp] 
%     \centering
%   \subfigure[$\epsilon=\infty$]{
%     \includegraphics[width=0.45\linewidth]{plots/new_c.png}
%     \label{fig:cp1} 
%   }
%   \subfigure[$\epsilon=5.5$]{
%     \includegraphics[width=0.45\linewidth]{plots/18_p55.png}
%     \label{fig:cp2} 
%   } 
%   \caption{ Images generated on CelebA dataset. The sub-figures show randomly generated faces (a) with and (b) without the private training procedure. In (b), a lot of faces are not recognizable, hence, preserving the privacy of such faces.}
%   \label{p1} 
% \end{figure*}

\subsection{E3: Results on CelebA dataset}

\begin{figure}[htbp] 
  \subfigure[Changing hair style]{
    \centering
    \includegraphics[width=0.95\linewidth]{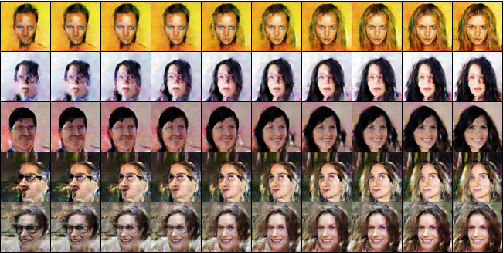}
    \label{fig:csp1} 
  }
  \subfigure[Changing brightness]{
    \centering
    \includegraphics[width=0.95\linewidth]{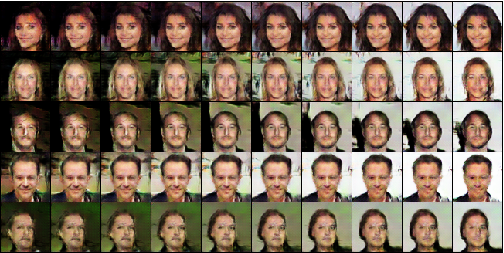}
    \label{fig:csp2} 
  }
  \caption{Varying latent codes on CelebA dataset. The images in both sub-figures are generated using a fixed, $\epsilon=5.5$. In both sub-figures, the discrete latent code is varied from top to bottom to generate a new face. In (a), the continuous latent code is varied from -1 to +1 (left to right) and shows variation in the hairstyle of generated images. In (b), the continuous latent code is varied from -1 to +1 (left to right) and shows variation in the brightness of generated images.}
  \label{csp} 
\end{figure}

We extended the experiments on CelebA dataset, which includes 202,599 celebrity face images with variations like pose and brightness. The experiment is performed several times with different values of $\epsilon$. We show the generated samples for $\epsilon=5.5$ as the generated images are lesser likely to be visually identifiable. As shown in Figure \ref{csp}, varying latent codes on CelebA dataset can generate images with varying faces, hair styles, and brightness. We used a uniform categorical latent code to capture variations in the face, and a uniform continuous latent code to capture style variations. Figure \ref{fig:csp1} shows the variation in continuous code brings change in the hairstyles of generated images. In Figure \ref{fig:csp2}, the continuous code captures the brightness of generated images, and varying it allows us to generate images with specific brightness.

\section{Discussion}
\label{disc}

imdpGAN framework can alleviate problems faced while working with imbalanced data by controlling the specificity of the generator output. For example, if the number of instances belonging to some class in a dataset is small, we can selectively generate samples (using the representative latent code) for that class to increase its samples. But the problem is, imdpGAN framework itself needs a sufficient amount of samples to learn the distributions before generating good quality data. Therefore, to solve the class imbalance problem using imdpGAN, we must have adequate samples belonging to the class having less number of samples in the dataset.

\begin{figure}[htbp]
    \centering
    \includegraphics[width=0.8\linewidth]{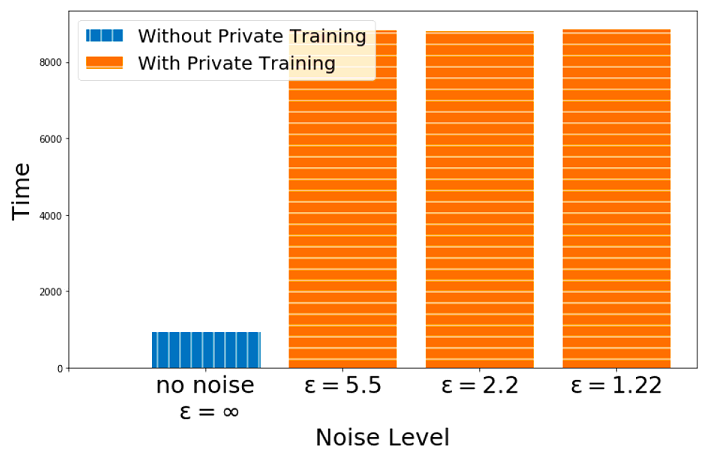}
    \caption{The time taken by the model with and without the private training procedure. Training time increases 10-fold with the private procedure. However, it is not affected by changing values of $\epsilon$.}
    \label{fig:time}
\end{figure}

\cite{Chen:2016:IIR:3157096.3157340} shows that mutual information regularization only adds a negligible complexity to GAN training. Therefore, to see how adding the private procedure effects the training time, we observed the training time without the private procedure, and with the private procedure for different noise levels. As shown in Figure \ref{fig:time}, the private training procedure introduces an approximately 10-fold increase in training time. Without the private procedure, the model takes 955 seconds for 50 epochs to train. On the other hand, with the private procedure added to the training, the same model takes 8,850 seconds for the same number of epochs. Note that the training time does not change much with varying noise levels ($\epsilon=\infty$ is the no noise case).

\section{Conclusion}
\label{conc}

In this paper, we have proposed an Information maximizing Differentially Private Generative Adversarial Network (imdpGAN), a unified framework to simultaneously preserve privacy and learn latent representations. imdpGAN preserves privacy and successfully learns meaningful latent representations. However, the private procedure added to imdpGAN's training results in a 10-fold increase in the training time. 

For future work, the core idea of using the private procedure and mutual information to learn latent representation can be applied to other datasets, which do not necessarily contain images. imdpGAN shows promising results on the image datasets: MNIST and CelebA. The architecture of the Generator and the Discriminator can be changed to extend imdpGAN to other data types.

\bibliographystyle{IEEEtran}
\bibliography{biblio}

\end{document}